%% file: naacl2021.tex
\pdfoutput=1

\documentclass[11pt]{article}

\usepackage[]{naacl2021}

\usepackage{times}
\usepackage{latexsym}

\usepackage[T1]{fontenc}

\usepackage[utf8]{inputenc}

\usepackage{microtype}
\usepackage{multirow}
\usepackage{tabularx}
\usepackage{colortbl}

\title{Synthetic Source Language Augmentation \\ for Colloquial Neural Machine Translation}

\author{Asrul Sani Ariesandy$^{\dagger\spadesuit}$ \And Mukhlis Amien$^{\dagger\clubsuit}$ \And Alham Fikri Aji$^{\spadesuit}$ \And Radityo Eko Prasojo$^{\spadesuit\diamondsuit}$ \AND
  {\normalfont $^\dagger$Sekolah Tinggi Informatika \& Komputer Indonesia (STIKI), Malang, Indonesia} \\ 
  $^\spadesuit$Kata.ai Research Team, Jakarta, Indonesia \\ $^\clubsuit$Beijing Institute of Technology, China $^\diamondsuit$Faculty of Computer Science, Universitas Indonesia \\ 
  $^\spadesuit$\texttt{\{asrul,aji,ridho\}@kata.ai} $^\dagger$\texttt{amien@stiki.ac.id}
  }

\begin{document}
\maketitle
\begin{abstract}

Neural machine translation (NMT) is typically domain-dependent and style-dependent, and it requires lots of training data. State-of-the-art NMT models often fall short in handling colloquial variations of its source language and the lack of parallel data in this regard is a challenging hurdle in systematically improving the existing models. In this work, we develop a novel colloquial Indonesian-English test-set collected from YouTube transcript and Twitter. We perform synthetic style augmentation to the source formal Indonesian language and show that it improves the baseline Id-En models (in BLEU) over the new test data. 

\end{abstract}

\input{content}

\bibliography{anthology,custom}
\bibliographystyle{acl_natbib}

\end{document}

%% file: content.tex
\section{Introduction}

Indonesian colloquialism is everyday and everywhere~\cite{wibowo2020semi}, e.g. in social media posts and conversational transcripts. Yet, existing research on Indonesian NLP models including NMTs often disregards qualitative analysis when the models are given strictly colloquial inputs~\cite{guntara-etal-2020-benchmarking}. This is mainly due to the fact that the data readily available for training and testing the models are in formal Indonesian. 

Colloquial Indonesian has several different word choices from formal language due to the diversity of regional languages and dialects. We define the spoken colloquial as a clean colloquial. In addition, in written media,  colloquial Indonesian is often abbreviated, disemvoweled, or written with voice alteration, which we define as the noisy colloquial (Example in Table~\ref{tab:contoh-col}).

\begin{table}[ht!]
   \small
    \centering
    \begin{tabular}{@{}ll@{}}
         \hline
         \textbf{Type} & \textbf{Example} \\
         \hline
         ~Formal      & \textcolor{blue}{Ayo$^1$} \textcolor{red}{bertemu$^2$} \textcolor{orange}{dengan$^3$} \textcolor{purple}{pak$^4$} Ridho \\
         ~Google Translate & Let's meet Pak Ridho \\ 
         ~Gold-standard & Let's meet Mr. Ridho \\ \rule{0pt}{3.5ex}
         Clean colloquial & \textcolor{blue}{kuy$^1$} \textcolor{red}{ketemu$^2$} \textcolor{orange}{sama$^3$} \textcolor{purple}{pak$^4$} Ridho \\
         ~Google Translate & kuy met Pak Ridho \\ \rule{0pt}{3.5ex}
         Noisy colloquial & \textcolor{blue}{kuy$^1$} \textcolor{red}{ktemu$^2$} \textcolor{orange}{sm$^3$} \textcolor{purple}{pk$^4$} ridho \\
         ~Google Translate & I met you at happy time \\
         \hline
    \end{tabular}
    \caption{Handcrafted example of colloquial Indonesian. Colloquial Indonesian used different different word (\textcolor{blue}{ayo-kuy$^1$} from Malang, \textcolor{red}{bertemu-ketemu$^2$}, \textcolor{orange}{dengan-sama$^3$} from Betawi). Social-media text introduces additional typographical noise, such as diemvowelling (\textcolor{red}{ketemu-ktemu$^2$}, \textcolor{orange}{sama-sm$^3$}, \textcolor{purple}{pak-pk$^4$}).}
    \label{tab:contoh-col}
\end{table}

To better evaluate English-Indonesian MT systems against colloquial text, we first create 2 new test-sets of Indonesian-English colloquial pairs. The first test is a clean colloquial taken from a YouTube transcript. The second test-set is a noisy colloquial from Twitter annotated by our team of annotators. We found that NMT systems trained on formal dataset did not perform very well on these test-sets.

Next, we develop synthetic colloquial text data by performing word-level translation of several words in the formal text into a colloquial form based on a word-to-word dictionary. By combining the formal dataset and the synthesized colloquial dataset, we increase the NMT performance on the colloquial test-set by 2.5 BLEU points.

\section{Related Work}

\citet{michel2018mtnt} developed a test-bed for noisy MT from social media. Related to that, \citet{vaibhav2019improving} showed that introducing synthetic noises improves NMT system evaluated on noisy test-set. This research draws inspiration from these works. However, the noise in colloquial Indonesian is different from the standard noise on social media, which is mostly typographical. In our case, the noise occurs in differences in the words used, because they are mixed with local languages or dialects, in addition of typographical noise.

Focusing on Indonesian-English MT itself, \citet{guntara-etal-2020-benchmarking} collected parallel English-Indonesian sentences in several domains, trained and benchmarked several NMT models, with some achieving state-of-the-art results. However, although the issue of colloquialism was briefly raised, there was no evaluation on model performance in translating either noisy or clean colloquial Indonesian to English. In this work, we make use of the training set they have provided.

On synthetic ``noise'' injection in Indonesian language, \citet{rizal-stymne-2020-evaluating} leveraged a dictionary-based approach to create a synthetic code-mixed data of Indonesian and English language and showed that the synthetized data is highly similar to an original code-mixed data. Code-mixing, especially with English, Arabic, and romanized Japanese or Korean, is a common type of colloqualism in Indonesian language.

Style transfer approaches have been attempted to standardize informal Indonesian~\cite{wibowo2020semi} with the motivation to serve as a preprocessing step for the downstream tasks (classification, sequence labeling, etc.), where the available models typically work well on formal Indonesian. High-quality pairs of informal-formal Indonesian data is again the main challenge in this work, and the authors pe  rformed iterative forward translation to generate more synthetic data, which was shown to improve the results up to some iterations.

\section{Low-Resource Colloquial-Indonesian to English}

The state-of-the-art model for Indonesian-English machine translation was trained and tested using formal Indonesian~\cite{guntara-etal-2020-benchmarking}, and thus its capability to translate colloquial Indonesian is untested. There is currently no available parallel colloquial Indonesian-English data either for testing or training. Therefore, we first manually create a high-quality test data to evaluate the existing models (Section~\ref{sec:test-corpus}).

Next, we investigate a way to improve the baseline results. Naturally, the most straightforward way is to add new training data in colloquial Indonesian, but it does not exist. As manual creation of translation data is time-consuming, we stop at manually creating test data, and for the training data we create synthetic data to augment the existing formal Indonesian dataset (Section~\ref{sec:synth-collo}).  

\subsection{Getting The Test Corpus}
\label{sec:test-corpus}

\begin{table*}[ht!]
\centering
    \begin{tabular}{ l c c c c c } 
     \hline
     \textbf{Source} & \textbf{Sentences} & \textbf{Total Words} & \textbf{OOV} & \textbf{AVG Token / Sent}  & \textbf{PPL} \\
     \hline
        Original Testset & 10.7K & 177.9K & 14.4\% & 16.5 & 115.8\\
        YouTube & 1.3K & 9.8K & 20.9\% & 6.4 & 120.3 \\
        Twitter & 2.0K & 25.2K & 33.6\% & 13.2 & 217.1 \\ 
     \hline
    \end{tabular}
    \caption{Exploratory data analysis of collected data. OOV stands for Out of Vocabulary words when compared against a standard dictionary~\cite{guntara-etal-2020-benchmarking}. AVG Token / Sent denotes average tokens per sentences.}
    \label{table:eda_for_collected}
\end{table*}

We first obtain the clean colloquial text data from Indonesian-language YouTube transcripts that contain English subtitles. Because it is a transcript of discussions and daily conversations, the Indonesian language used is colloquial. We get our transcripts from Indonesian channels consisting of Hotman Paris\footnote{\url{https://www.youtube.com/channel/UCU_fuBEbZ0Wcf3ZNWEtxwzg}} and TaulanyTV.\footnote{\url{https://www.youtube.com/channel/UC6SPCnTAIanF2_8ST2wrQzw}} We then filtered the transcript data by removing sentences that are less than 5 tokens. Our final test set for this scenario consists of 1317 sentences.

Then, we get the noisy colloquial text data from Twitter. We scrape tweets from users spread across 34 provinces in Indonesia from 2015 to 2020 (captured by the tweet's geolocation). We also do the same filtering process as we did on YouTube Transcript. Then we sampled and translated 2000 tweets into English by the annotators.
To show that this new informal dataset is really different than the existing formal Indonesian, we check the OOV-rate and the perplexity of the dataset. The OOV-rate is checked against the vocabulary used in Indonesian Wikipedia, which uses formal language. The perplexity is computed with GPT-2 casual language modelling trained on Wikipedia.\footnote{\url{https://github.com/cahya-wirawan/indonesian-language-models/tree/master/ULMFiT}} These results can be seen on Table~\ref{table:eda_for_collected}.

\subsection{Generating Synthetic Colloquial}
\label{sec:synth-collo}

Developing colloquial Indonesian-English for training is expensive. Therefore, we opt to create it synthetically by performing word-level formal to informal transformations on the original dataset using a colloquial dictionary.\footnote{\url{https://github.com/louisowen6/NLP_bahasa_resources}}

We create synthetic colloquial data from both sets by randomly changing the formal Indonesian words contained in the dictionary into informal words. For each word, we ``roll a dice'' that is normally distributed from 0 to 1. If the dice rolls under a certain threshold, then the word is changed into its informal version, and vice versa. We used 4 different thresholds, 0.3, 0.5, 0.7 and 1.0, to generate 4 different datasets (Table~\ref{table:eda_for_generated}). This means that the bigger the threshold, the more informal words are being introduced into the dataset. Generally, a formal word might have be changed into several colloquial words. In that case, the informal word will be randomly selected with uniform probability. The English side of the modified data is left intact. 

\begin{table}[ht!]
\centering
    \begin{tabular}{l c} 
     \hline
     \textbf{Source} & \textbf{Transformed\%}  \\
     \hline
        Colloquial 0\%$^*$ & 0\%  \\
        Colloquial 30\% & 7.1\%  \\
        Colloquial 50\% & 11.9\%  \\
        Colloquial 70\% & 16.6\%  \\
        Colloquial 100\% & 23.8\%  \\
     \hline
    \end{tabular}
    \caption{Exploratory data analysis of generated data. Transformed denotes total colloquialized words. Transformed\% denotes the actual percentage of transformed words. $^*$0\% colloquial is equal to the original dataset.}
    \label{table:eda_for_generated}
\end{table}

Since the dictionary only covers a limited number of words, we evaluate our synthetic dataset in terms of the actual formal words that are being transformed.

\section{Experimental Setting}

Our NMT system uses the standard Transformer architecture~\cite{google_att} of 6 encoder layers and 6 decoder layers with tied embedding.  Our models are trained up to a maximum of 5 epochs, or until we observe no improvement for 10 consecutive validation steps. The rest of the hyperparameters follows the suggested value~\cite{google_att}.

Before being applied to the training process, we made several transformations to our datasets. To improve the model's quality, we also use additional 4.7M back-translated~\cite{rico_bt} data from the OSCAR corpus~\cite{ortiz-suarez-etal-2020-monolingual}. We applied byte-pair encoding~\cite{subword_nmt} with 30k merge operations to preprocess the input. We use the Marian toolkit to train our model~\cite{junczys2018marian}.

\section{Experiment Result}

Our experiment is separated into five scenarios as mentioned in Table~\ref{table:eda_for_generated}. We start with a baseline model that is only trained with the formal Indonesian-English parallel data \cite{guntara-etal-2020-benchmarking} and back-translation data \cite{ortiz-suarez-etal-2020-monolingual}. We then add the synthetic colloquial dataset on top of that with different colloquial ratio. 
The results are shown in Table~\ref{table:bleu_result}. We measured our model results with BLEU \cite{papineni-etal-2002-bleu}, computed using sacreBLEU~\cite{post2018call}.

\begin{table*}[ht!]

\centering
    \begin{tabular}{l c c c c } 
     \hline
      \multirow{2}{*}{\textbf{Training Data}} & \multicolumn{4}{c}{\textbf{Evaluation BLEU}} \\  \cline{2-5}
      & \textbf{Original} & \textbf{YouTube} & \textbf{Twitter} & \textbf{Avg.} \\ [0.5ex]
     \hline
        Original & 29.4 & 31.6 & 12.6 & 24.53 \\
        Original + Synth. Colloquial 30\% & 29.0 & 32.1 & 13.5 & 24.87 \\
        Original + Synth. Colloquial 50\% & 28.9 & 32.5 & 13.2 & 24.87 \\
        Original + Synth. Colloquial 70\% & \textbf{29.4} & \textbf{34.1} & \textbf{13.6} & \textbf{25.70} \\
        Original + Synth. Colloquial 100\% & 29.0 & 34.0 & 13.5 & 25.50 \\
     \hline
    \end{tabular}
    \caption{BLEU Scores result for each test sets. Original data denotes training set from \citet{guntara-etal-2020-benchmarking} + back-translated OSCAR.}
    \label{table:bleu_result}
\end{table*}

From the results presented in Table~\ref{table:bleu_result}, it shows that there has been an increase in translation results after we add the augmentation data. The highest increase was shown by Colloquial 70\% data with  average BLEU result of 25.70. While the lowest increase was generated by the Colloquial 30\% and Colloquial 50\% data. 

\section{Discussion}

Our results show that adding synthetic data improves the overall BLEU and therefore this is a potential direction for further research. The results shown in Table~\ref{table:bleu_result} suggest that the sweet spot of 70\% is due to a ``correct'' amount of proportion of colloquialism injected to the original data. However, it might be the case that some certain types of colloquialism are more important than the other, and this affects the overall results due to the randomness and incompleteness (due to the dictionary limitation) of our transformation. Injecting more relevant colloquialism, while avoiding the less relevant ones, may potentially increase the overall model performance.

Our models performed poorly on the Twitter test-set, with minimal improvements when using synthetic datasets. This result shows that noisy colloquial Indonesian is challenging to handle, as the language is extremely different from formal Indonesian. Moreover, typographical noises are quite common, adding extra difficulty. These type of noises are not handled by simply switching some words into their colloquial form. We argue that handling noisy colloquial Indonesian requires synthetic data beyond word-to-word transformation. Empirically, Table~\ref{table:eda_for_generated} also shows the limitation of our dictionary-based word-to-word transformation, as indicated by the Colloquial 100\% dataset having only 23.8\% transformed tokens, while an average Indonesian tweet has 33\% OOV words in it. In the future, we plan to train a model to generate the synthetic data, making use of the model and data provided by \citet{wibowo2020semi}, to be able to generate colloquialism even for unseen words or phrases.

\section{Conclusion}

We investigated data augmentation methods to improve NMT's translation capabilities by converting formal Indonesian data to colloquial and adding he experimental results it to our dataset. With several test schemes show an increase in the translation ability of the models that have been made. Determining how many words to covert into colloquial Indonesian is important. To the best of our knowledge, we are the first to perform a data augmentation method for Colloquial Indonesian at NMT.

\section*{Acknowledgement}

We would like to thank Haryo Akbarianto Wibowo and Made Nindyatama Nityasya for the fruitful discussions. We also would like to thank our linguistic data annotators: Suci Fitriany, Salma Qonitah, Tomi Santoso, and Nadhifa Zulfa for providing some of the data used in this work.

%% file: naacl2021.bbl
\begin{thebibliography}{12}
\expandafter\ifx\csname natexlab\endcsname\relax\def\natexlab#1{#1}\fi

\bibitem[{Guntara et~al.(2020)Guntara, Aji, and
  Prasojo}]{guntara-etal-2020-benchmarking}
Tri~Wahyu Guntara, Alham~Fikri Aji, and Radityo~Eko Prasojo. 2020.
\newblock \href {https://www.aclweb.org/anthology/2020.bucc-1.6} {Benchmarking
  multidomain {E}nglish-{I}ndonesian machine translation}.
\newblock In \emph{Proceedings of the 13th Workshop on Building and Using
  Comparable Corpora}, pages 35--43, Marseille, France. European Language
  Resources Association.

\bibitem[{Junczys-Dowmunt et~al.(2018)Junczys-Dowmunt, Grundkiewicz, Dwojak,
  Hoang, Heafield, Neckermann, Seide, Germann, Fikri~Aji, Bogoychev, Martins,
  and Birch}]{junczys2018marian}
Marcin Junczys-Dowmunt, Roman Grundkiewicz, Tomasz Dwojak, Hieu Hoang, Kenneth
  Heafield, Tom Neckermann, Frank Seide, Ulrich Germann, Alham Fikri~Aji,
  Nikolay Bogoychev, Andr\'{e} F.~T. Martins, and Alexandra Birch. 2018.
\newblock \href {http://www.aclweb.org/anthology/P18-4020} {Marian: Fast neural
  machine translation in {C++}}.
\newblock In \emph{Proceedings of ACL 2018, System Demonstrations}, pages
  116--121, Melbourne, Australia. Association for Computational Linguistics.

\bibitem[{Michel and Neubig(2018)}]{michel2018mtnt}
Paul Michel and Graham Neubig. 2018.
\newblock Mtnt: A testbed for machine translation of noisy text.
\newblock In \emph{Proceedings of the 2018 Conference on Empirical Methods in
  Natural Language Processing}, pages 543--553.

\bibitem[{Ortiz~Su{\'a}rez et~al.(2020)Ortiz~Su{\'a}rez, Romary, and
  Sagot}]{ortiz-suarez-etal-2020-monolingual}
Pedro~Javier Ortiz~Su{\'a}rez, Laurent Romary, and Beno{\^\i}t Sagot. 2020.
\newblock \href {https://www.aclweb.org/anthology/2020.acl-main.156} {A
  monolingual approach to contextualized word embeddings for mid-resource
  languages}.
\newblock In \emph{Proceedings of the 58th Annual Meeting of the Association
  for Computational Linguistics}, pages 1703--1714, Online. Association for
  Computational Linguistics.

\bibitem[{Papineni et~al.(2002)Papineni, Roukos, Ward, and
  Zhu}]{papineni-etal-2002-bleu}
Kishore Papineni, Salim Roukos, Todd Ward, and Wei-Jing Zhu. 2002.
\newblock \href {https://doi.org/10.3115/1073083.1073135} {{B}leu: a method for
  automatic evaluation of machine translation}.
\newblock In \emph{Proceedings of the 40th Annual Meeting of the Association
  for Computational Linguistics}, pages 311--318, Philadelphia, Pennsylvania,
  USA. Association for Computational Linguistics.

\bibitem[{Post(2018)}]{post2018call}
Matt Post. 2018.
\newblock A call for clarity in reporting bleu scores.
\newblock In \emph{Proceedings of the Third Conference on Machine Translation:
  Research Papers}, pages 186--191.

\bibitem[{Rizal and Stymne(2020)}]{rizal-stymne-2020-evaluating}
Arra{'}Di~Nur Rizal and Sara Stymne. 2020.
\newblock \href {https://www.aclweb.org/anthology/2020.calcs-1.4} {Evaluating
  word embeddings for {I}ndonesian{--}{E}nglish code-mixed text based on
  synthetic data}.
\newblock In \emph{Proceedings of the The 4th Workshop on Computational
  Approaches to Code Switching}, pages 26--35, Marseille, France. European
  Language Resources Association.

\bibitem[{Sennrich et~al.(2016{\natexlab{a}})Sennrich, Haddow, and
  Birch}]{rico_bt}
Rico Sennrich, Barry Haddow, and Alexandra Birch. 2016{\natexlab{a}}.
\newblock \href {https://doi.org/10.18653/v1/P16-1009} {Improving neural
  machine translation models with monolingual data}.
\newblock In \emph{Proceedings of the 54th Annual Meeting of the Association
  for Computational Linguistics (Volume 1: Long Papers)}, pages 86--96, Berlin,
  Germany. Association for Computational Linguistics.

\bibitem[{Sennrich et~al.(2016{\natexlab{b}})Sennrich, Haddow, and
  Birch}]{subword_nmt}
Rico Sennrich, Barry Haddow, and Alexandra Birch. 2016{\natexlab{b}}.
\newblock \href {https://doi.org/10.18653/v1/P16-1162} {Neural machine
  translation of rare words with subword units}.
\newblock In \emph{Proceedings of the 54th Annual Meeting of the Association
  for Computational Linguistics (Volume 1: Long Papers)}, pages 1715--1725,
  Berlin, Germany. Association for Computational Linguistics.

\bibitem[{Vaibhav et~al.(2019)Vaibhav, Singh, Stewart, and
  Neubig}]{vaibhav2019improving}
Vaibhav Vaibhav, Sumeet Singh, Craig Stewart, and Graham Neubig. 2019.
\newblock Improving robustness of machine translation with synthetic noise.
\newblock In \emph{Proceedings of the 2019 Conference of the North American
  Chapter of the Association for Computational Linguistics: Human Language
  Technologies, Volume 1 (Long and Short Papers)}, pages 1916--1920.

\bibitem[{Vaswani et~al.(2017)Vaswani, Shazeer, Parmar, Uszkoreit, Jones,
  Gomez, Kaiser, and Polosukhin}]{google_att}
Ashish Vaswani, Noam Shazeer, Niki Parmar, Jakob Uszkoreit, Llion Jones,
  Aidan~N Gomez, \L~ukasz Kaiser, and Illia Polosukhin. 2017.
\newblock \href
  {http://papers.nips.cc/paper/7181-attention-is-all-you-need.pdf} {Attention
  is all you need}.
\newblock In \emph{Advances in Neural Information Processing Systems 30}, pages
  5998--6008. Curran Associates, Inc.

\bibitem[{Wibowo et~al.(2020)Wibowo, Prawiro, Ihsan, Aji, Prasojo, Mahendra,
  and Fitriany}]{wibowo2020semi}
Haryo~Akbarianto Wibowo, Tatag~Aziz Prawiro, Muhammad Ihsan, Alham~Fikri Aji,
  Radityo~Eko Prasojo, Rahmad Mahendra, and Suci Fitriany. 2020.
\newblock Semi-supervised low-resource style transfer of indonesian informal to
  formal language with iterative forward-translation.
\newblock In \emph{2020 International Conference on Asian Language Processing
  (IALP)}. IEEE.

\end{thebibliography}
